# IMPROVED SPOKEN DOCUMENT SUMMARIZATION WITH COVERAGE MODELING TECHNIQUES

*Kuan-Yu Chen\*, Shih-Hung Liu\*, Berlin Chen†, Hsin-Min Wang\**

\*Academia Sinica, Taipei, Taiwan
†National Taiwan Normal University, Taipei, Taiwan

\*{kychen, journey, whm}@iis.sinica.edu.tw, †berlin@ntnu.edu.tw

## ABSTRACT

Extractive summarization aims at selecting a set of indicative sentences from a source document as a summary that can express the major theme of the document. A general consensus on extractive summarization is that both relevance and coverage are critical issues to address. The existing methods designed to model coverage can be characterized by either reducing redundancy or increasing diversity in the summary. Maximal margin relevance (MMR) is a widely-cited method since it takes both relevance and redundancy into account when generating a summary for a given document. In addition to MMR, there is only a dearth of research concentrating on reducing redundancy or increasing diversity for the spoken document summarization task, as far as we are aware. Motivated by these observations, two major contributions are presented in this paper. First, in contrast to MMR, which considers coverage by reducing redundancy, we propose two novel coverage-based methods, which directly increase diversity. With the proposed methods, a set of representative sentences, which not only are relevant to the given document but also cover most of the important sub-themes of the document, can be selected automatically. Second, we make a step forward to plug in several document/sentence representation methods into the proposed framework to further enhance the summarization performance. A series of empirical evaluations demonstrate the effectiveness of our proposed methods.

## 1. INTRODUCTION

With the rapid development of the Internet, exponentially growing multimedia content, such as music video, broadcast news programs, and lecture recordings, has been continuously filling our daily life [1-4]. The overwhelming data inevitably leads to an information overload problem. Since speech is one of the most important sources of information in the multimedia content, by virtue of spoken document summarization (SDS), one can efficiently digest or browse the multimedia content by listening to the associated speech summary. Extractive SDS, which manages to select a set of indicative sentences from a spoken document according to a target summarization ratio and concatenate them to form a summary, has thus been an attractive research topic in recent years [5-8].

A general consensus on extractive summarization is that both *relevance* and *coverage* are critical issues in a realistic scenario [9-13]. However, most of the existing summarization methods focus on determining only the relevance degree between a given document and one of its sentences [14-18]. As a result, the top-ranked sentences returned by these methods may only cover partial sub-themes of the given document and fail to interpret the whole picture. Summarization result diversification is devoted to covering important aspects (or sub-themes) of a document as many as possible. The developed methods following this line of research on coverage modeling can be categorized into *implicit* and *explicit* methods [19-21]. Formally, an implicit method reduces redundancy in a summary by considering sentence similarities, while an explicit method increases diversity of a summary by taking the sub-themes of the document into consideration.

Maximal margin relevance (MMR) [10], which iteratively selects a sentence that has the highest combination of a similarity score with respect to the given document and a dissimilarity score with respect to those already selected sentences, is a canonical representative of the implicit methods. However, aside from MMR, there is still little focus on investigating summarization result diversification. In view of this, we propose two novel coverage-based methods for extractive spoken document summarization. By leveraging these methods, a concise summary can be automatically generated by rendering not only relevance but also coverage. We also explore to incorporate several document/sentence representation methods into the proposed framework to further enhance the summarization performance.

## 2. RELATED WORK

The wide spectrum of extractive SDS methods developed so far spreads from methods simply based on the sentence position or structure information, methods based on unsupervised sentence ranking, to methods based on supervised sentence classification [5, 8]. For the first category, important sentences are selected from some salient parts of a spoken document [10], such as the introductory and/or concluding parts. However, such methods can be only applied to some specific domains with limited document structures. Unsupervised sentence ranking methods attempt to select important sentences based on some statistical features of the sentences or of the words in the sentences without human annotation involved. Popular methods include, but are not limited to, vector space model [22], latent semantic analysis [22], Markov random walk [23], MMR [10], sentence significant score method [24], language model-based framework [17, 18], LexRank [25], submodularity-based method (SM) [26], and integer linear programming-based method (ILP) [27]. The statistical features may include, for example, the term (word) frequency, linguistic score, recognition confidence measure, and prosodic information. In contrast, supervised sentence classification methods, such as Gaussian mixture model [28], Bayesian classifier [29], support vector machines (SVM) [30, 31], and conditional random fields (CRF) [32], usually formulate sentence selection as a binary classification problem, i.e., a sentence can either be included in a summary or not. Interested readers may refer to [5-8] for comprehensive reviews and new insights into the major methods that have been developed and applied with good success to a wide range of text and spoken document summarization tasks.

In addition to MMR, the ability of reducing redundancy (or increasing diversity) has also been aptly incorporated into SM, ILP, and the structured SVM method [31]. However, SM and ILP are not readily suited for large-scale problems, since they involve a rather time-consuming process in important sentence selection. On the other hand, the structured SVM method needs a set of training documents along with their corresponding handcrafted summaries, which is difficult to collect because manual annotation is both time-consuming and labor-intensive, for training the classifiers (or summarizers). In view of this, we are intended to develop an unsupervised summarization framework that can simultaneously take both relevance and coverage into account in a principled and effective manner.

## 3. COVERAGE MODELING TECHNIQUES

Perhaps the most common belief in the document summarization community is that relevance and coverage are two key issues for generating a concise summary. For the idea to go, a principled realization of progressively selecting important sentences can be formulated as [10, 19]

$$S^* = \arg\max_{S \in D-\mathbf{S}} Rel(D,S) + \alpha \cdot Cov(D,\mathbf{S},S), \quad (1)$$

where $D$ denotes a given document to be summarized, $\mathbf{S}$ is a set of sentences that have already been selected, and $S$ is one of the candidate sentences in $D$. $Rel(\cdot,\cdot)$ is a similarity function used to determine the relevance degree between the source document and one of its sentences, and $Cov(\cdot,\cdot,\cdot)$ denotes a coverage function. In the context of MMR, the coverage score for a candidate sentence may be computed by [10]

$$Cov_{\text{MMR}}(D,\mathbf{S},S) \equiv -\frac{1}{|\mathbf{S}|}\sum_{S' \in \mathbf{S}} Rel(S',S). \quad (2)$$

Intuitively, MMR iteratively selects a sentence that is not only relevant to the document but also dissimilar to the already selected sentences.

### 3.1. The xDTD Method

As opposed to MMR, which reduces redundancy by using similarities between summary sentences, another promising direction to consider coverage is to increase the diversity of a summary. Formally, given a document $D$, the probability that a sentence $S$ meets the gold summary $\mathbf{S}^*$ can be written as

$$P(\mathbf{S}^* | D,S) = \frac{P(S | \mathbf{S}^*,D)P(\mathbf{S}^*,D)}{P(D,S)} \propto P(S | \mathbf{S}^*,D). \quad (3)$$

Since $P(\mathbf{S}^*,D)$ will not affect the ranking of a sentence and the prior probability of a sentence can be assumed identical for all sentences in the document, we can thus omit $P(\mathbf{S}^*,D)$ and $P(D,S)$ in Eq. (3) and evaluate the sentence by $P(S|\mathbf{S}^*,D)$.

Next, it is reasonable that the gold summary covers all the important sub-themes of the document. Therefore, by taking sub-themes of the document into consideration, we obtain

$$P(S | \mathbf{S}^*,D) = \sum_{k=1}^{K} P(S | T_k) P(T_k | \mathbf{S}^*,D), \quad (4)$$

where $T_k$ is the $k$-th sub-theme in $D$; $P(S|T_k)$ stands for the coverage degree of sentence $S$ under the $k$-th sub-theme; and $P(T_k|\mathbf{S}^*,D)$ can be seen as a relative importance measure of the sub-theme $T_k$, subject to $\sum_{k=1}^{K} P(T_k|\mathbf{S}^*,D) = 1$. Although the gold summary $\mathbf{S}^*$ cannot be obtained at the test stage, it is generally agreed upon that a concise summary for a document should cover most of the important aspects of the document. Consequently, the coverage score can further be simplified as

$$Cov_{\text{xDTD}}(D,\mathbf{S},S) \equiv \sum_{k=1}^{K} P(S|T_k) P(T_k | D). \quad (5)$$

We name the model eXplicit Document sub-Theme Diversification (xDTD for short hereafter).

In our practical implementation, $P(S|T_k)$ is computed by

$$P(S | T_k) = \frac{Rel(S,T_k)}{\sum_{S' \in D} Rel(S',T_k)}, \quad (6)$$

and $P(T_k|D)$ is estimated in a similar manner.

### 3.2. The J-xDTD Method

On top of MMR and xDTD, which implicitly and explicitly model coverage by considering redundancy and diversity, respectively, a more comprehensive method can be proposed. As an extension from $P(S|\mathbf{S}^*,D)$, we hence define the coverage score as

$$Cov_{\text{J-xDTD}}(D,\mathbf{S},S) \equiv P(S,\overline{\mathbf{S}} | \mathbf{S}^*,D), \quad (7)$$

which interprets the likelihood of observing a candidate sentence $S$ but ***not*** those already selected sentences (denoted as $\overline{\mathbf{S}}$). Again, by explicitly considering the sub-themes inherent in the given document $D$, the likelihood can be decomposed as

$$P(S,\overline{\mathbf{S}} | \mathbf{S}^*,D) = \sum_{k=1}^{K} P(S,\overline{\mathbf{S}} | T_k) P(T_k | \mathbf{S}^*,D). \quad (8)$$

Next, by assuming that $S$ and $\overline{\mathbf{S}}$ are conditionally independent given a sub-theme, we obtain

$$P(S,\overline{\mathbf{S}} | T_k) = P(S | T_k) P(\overline{\mathbf{S}} | T_k). \quad (9)$$

Obviously, the former term (i.e., $P(S|T_k)$) is used to model the coverage of sentence $S$ with respect to each sub-theme $T_k$, and the latter provides a novelty measure to determine the ***dissatisfaction*** degree of the sub-theme $T_k$ for those already selected sentences. By assuming that sentences in $\mathbf{S}$ are independent given a sub-theme, we can estimate the dissatisfied degree by

$$P(\overline{\mathbf{S}} | T_k) = \prod_{S' \in \mathbf{S}} (1 - P(S' | T_k)). \quad (10)$$

Since the method extends the concept of xDTD by jointly taking redundancy and diversity into consideration, we refer to it as "J-xDTD" hereafter.

### 3.3. Analytic Comparisons & Implementation Details

Several analytic comparisons can be made among the aforementioned three coverage modeling techniques. First, the coverage-based methods can be characterized by either reducing redundancy or increasing diversity. MMR belongs to the first category and xDTD can be classified into the second category, while J-xDTD takes both redundancy and diversity into account simultaneously. On one hand, a marked difference between MMR and J-xDTD is that the former compares a sentence to every already selected sentence, whereas the latter leverages $\mathbf{S}$ to estimate the dissatisfied degree for each sub-theme at each sentence selection iteration. On the other hand, the major distinction between the proposed two coverage-based methods is that xDTD determines the importance degree for each sub-theme by only referring to the document itself while J-xDTD considers both those already selected sentences and the document. To sum up, the importance degree for each sub-theme is dynamically determined at each sentence selection iteration for J-xDTD, but is kept fixed during the sentence selection process for xDTD. Next, both MMR and J-xDTD select the indicative sentences in a recursive manner, while xDTD generates a summary through a one-pass process. Thus, in practical implementation, MMR and J-xDTD are slightly slower than xDTD. Moreover, xDTD and J-xDTD have their roots in the information retrieval community [19-21]. This is the first time that xDTD and J-xDTD are formally introduced, adapted, and evaluated in the SDS task, as far as we are aware.

Noticeably, sub-themes play a fundamental role within the proposed coverage-based methods (*cf.* Sections 3.1 and 3.2). However, in reality, the syntactic/semantic sub-themes of a document are hard to determine. As a pilot study on empirical comparison of coverage-based methods, in this paper, we treat each sentence in a document as a sub-theme. The similarity function (i.e., $Rel(\cdot,\cdot)$) involved in MMR and the proposed methods is estimated based on the cosine similarity measure. We normalize the document/sentence/sub-theme representations (*cf.* Section 4) to unit vectors to speed up the calculation and make the resulting similarity scores range between 0 and 1.

## 4. DOCUMENT/SENTENCE REPRESENTATIONS

### 4.1. Bag-of-Words Representation

The bag-of-words (BOW) representation has long been a basis for most of the natural language processing-related tasks. The major advantage of BOW is that it is not only simple and intuitive, but also efficient and effective. In BOW, each document/sentence is represented by a high-dimensional vector, where each dimension specifies the occurrence statistics associated with an index term (e.g., word, subword, or their *n*-grams) in the document/sentence. To eliminate the noisy words (e.g., the function words) and promote the discriminative words (e.g., the content words), the statistics is usually estimated with the term frequency (TF) weighted by the inverse document frequency (IDF) [7].

### 4.2. Distributed Representation

On the other hand, representation learning has emerged as a newly favorite research subject because of its excellent performance [33, 34]. However, as far as we are aware, there are relatively few studies investigating its use in extractive text or spoken document summarization. Well-known methods for document/sentence embedding include the distributed memory (DM) model [35] and the distributed bag-of-words (DBOW) model [35, 36], to name just a few.

#### 4.2.1 *The Distributed Memory Model*

The DM model is inspired and hybridized by the traditional feed-forward neural network language model (NNLM) [37]. Formally, based on the NNLM, the idea underlying the DM model is that a given paragraph and a predefined number of context words should jointly contribute to the prediction of the next word [35]. To this end, the objective function is defined as

$$\sum_{i=1}^{|\mathbf{D}|} \sum_{j=1}^{|D_i|} \log P(w_j \mid w_{j-n+1},\ldots,w_{j-1}, D_i), \quad (11)$$

where $|\mathbf{D}|$ is the number of paragraphs in the training corpus $\mathbf{D}$, $D_i$ denotes the *i*-th paragraph, and $|D_i|$ is the length of $D_i$. Since the model acts as a memory unit that remembers what is missing from the current context, it is named the distributed memory (DM) model.

In our implementation, given a document, each sentence in the document and the document itself are considered as a paragraph (i.e., $D_i$), and the vector representations of the document and all its sentences are obtained by maximizing the objective function depicted in Eq. (11).

#### 4.2.2. *The Distributed Bag-of-Words Model*

A simplified version of the DM model is to merely draw on the paragraph representations to predict all of the words in the paragraphs [30]. The objective function is then defined as

$$\sum_{i=1}^{|\mathbf{D}|} \sum_{j=1}^{|D_i|} \log P(w_j \mid D_i). \quad (12)$$

Since the simplified model ignores the contextual words at the input layer, it is named the distributed bag-of-words (DBOW) model. The document/sentence representations can be obtained in a similar manner as the DM model.

## 5. EXPERIMENTAL SETUP

The dataset used in this study is the MATBN broadcast news corpus collected by the Academia Sinica and the Public Television Service Foundation of Taiwan between November 2001 and April 2003 [38]. The corpus has been segmented into separate stories and transcribed manually. Each story contains the speech of one studio anchor, as well as several field reporters and interviewees. A subset of 205 broadcast news documents compiled between November 2001 and August 2002 was reserved for the summarization experiments. We chose 20 documents as the test set while the remaining 185 documents as the held-out development set. The reference summaries were generated by ranking the sentences in the manual transcript of a spoken document by importance without assigning a score to each sentence. Each document has three reference summaries annotated by three subjects. For the assessment of summarization performance, we adopted the widely-used ROUGE metrics [39]. All the experimental results reported hereafter are obtained by calculating the F-scores [30] of these ROUGE metrics. The summarization ratio was set to 10%. A subset of 25-hour speech data from MATBN compiled from November 2001 to December 2002 was used to bootstrap acoustic model training with a minimum phone error rate (MPE) criterion and a training data selection scheme [40]. The vocabulary size is about 72 thousand words. The average word error rate of automatic transcription is about 40%.

## 6. EXPERIMENTAL RESULTS

In the first set of experiments, we evaluate the utilities of different paragraph embedding methods (i.e., BOW, DM, and DBOW) for document/sentence representation in extractive summarization task. Sentences in a given document to be summarized are ranked solely by the similarity degree between each sentence and the document, and in turn be selected to form the final summery. The results are shown in Table 1, where TD denotes the results obtained based on the manual transcripts of spoken documents and SD denotes the results using the speech recognition transcripts that may contain recognition errors. From the results, several observations can be made. First, although BOW is a simple and intuitive representation method, it outperforms DM and DBOW in both the TD and SD cases. Second, DBOW outperforms DM in both cases, although the former is a simplified variant of the latter. Third, although the simple and efficient ability of BOW has been evidenced, an obvious shortcoming of BOW is that it cannot address synonymy and polysemy words well. As such, simply matching words occurring in a sentence and a document may not capture the semantic intent within them. Distributed representation methods are capable of mitigating the difficulty to some extent. An intuitive strategy is to concatenate both types of representations together. The experimental results are also shown in Table 1 (*cf.* BOW+DM and BOW+DBOW). As expected, the combinative representations outperform their respective component representation methods by a large margin in both the TD and SD cases. An interesting observation is that the performance gap between DM and DBOW seems to be reduced when they are combined with BOW. Accordingly, we will incorporate BOW, BOW+DM, and BOW+DBOW with MMR and the proposed coverage-based methods, respectively, in the following experiments.

In the next set of experiments, we evaluate the proposed coverage-based methods (i.e., xDTD and J-xDTD). The celebrated MMR method, which considers both relevance and redundancy when generating a summary, is treated as the baseline system. The results are shown in Table 2. From the viewpoint of the representation method, when pairing with BOW+DM, both xDTD and J-xDTD perform quite well, while BOW+DBOW seems to be better suited for MMR. The combinative representation methods (i.e., BOW+DM and BOW+DBOW) outperform the BOW method again, when in conjugated with the enhanced summarization methods. It seems that the summarization results cannot be further improved when BOW+DBOW is incorporated with the coverage-based methods (i.e., MMR, xDTD, and J-xDTD) in both the TD and SD cases. The reason should be further studied. Lastly, when compared with the baseline MMR system, the proposed methods demonstrate their superiority in the TD case, while they only achieve comparable results with MMR in the SD case. A possible reason might be that imperfect speech recognition may drift the estimation for the sub-themes of each document. Thus, xDTD and J-xDTD may not benefit from taking sub-themes into account. However, the results still confirm the capabilities of the proposed methods in the TD case, especially when pairing with BOW+DM.

In the last set of experiments, we assess the performance levels of several well-practiced or/and state-of-the-art summarization methods for extractive summarization, including the variations of the vector-space model (i.e., latent semantic analysis (LSA), continuous bag-of-words model (CBOW), skip-gram model (SG), and global vector model (GloVe)), the language model-based summarization method (i.e., unigram language model (ULM)), the graph-based methods (i.e., Markov random walk (MRW) and LexRank), and the combinatorial optimization methods (i.e., SM and ILP). The results are presented in Table 3. Several noteworthy observations can be drawn from the table. First, LSA, which represents the sentences of a spoken document and the document itself in the latent semantic space instead of the index term (word) space, performs slightly better than BOW in both the TD and SD cases (*cf.* Table 1). Next, the three word embedding methods (i.e., CBOW, SG, and GloVe), though with disparate model structures and learning strategies, achieve comparable results to one another in both the TD and SD cases. Note here that they are also concatenated with the BOW representation method in our implementation. An interesting comparison is that BOW+DM and BOW+DBOW outperform them as expected in the TD case, but offer only a small performance gain in the SD case (*cf.* Table 1). Third, the two graph-based methods (i.e., MRW and LexRank) are quite competitive with each other and perform better than the vector-space methods (i.e., LSA, CBOW, SG, and GloVe) in the TD case. However, in the SD case, the situation is reversed. It reveals that imperfect speech recognition may negatively affect the graph-based methods more than the vector-space methods; a possible reason for such a phenomenon is that the speech recognition errors may lead to inaccurate similarity measures between each pair of sentences. The PageRank-like procedure of the graph-based methods, in turn, will be performed based on these problematic measures, potentially leading to degraded results. Fourth, ULM shows results comparable to other state-of-the-art methods in both the TD and SD cases. Finally, SM and ILP stand out in performance in the TD case, but only deliver results on par with the other methods in the SD case. When pairing with BOW+DM, the proposed methods can achieve comparable results with the combinatorial optimization methods in the TD case and outperform them in the SD case (*cf.* Tables 2 & 3). Although, both SM and ILP aptly integrate the ability of reducing redundancy (or increasing diversity) for summarization, they are heavyweight methods (*cf.* Section 2). Thus, the results support the potential of the proposed methods in practical applications.

**Table 1.** *Summarization results achieved by document/sentence representations with different paragraph embedding methods.*

| Method | Text Documents (TD) | | | Spoken Documents (SD) | | |
|---|---|---|---|---|---|---|
| | Rouge-1 | Rouge-2 | Rouge-L | Rouge-1 | Rouge-2 | Rouge-L |
| BOW | 0.347 | 0.228 | 0.290 | 0.342 | 0.189 | 0.287 |
| DM | 0.301 | 0.174 | 0.264 | 0.264 | 0.118 | 0.226 |
| DBOW | 0.322 | 0.215 | 0.289 | 0.292 | 0.152 | 0.258 |
| BOW+DM | 0.406 | 0.290 | 0.355 | 0.364 | 0.218 | 0.313 |
| BOW+DBOW | 0.418 | 0.293 | 0.364 | 0.375 | 0.232 | 0.323 |

**Table 2.** *Summarization results achieved by the proposed summarization framework with different representation methods.*

| Method | | Text Documents (TD) | | | Spoken Documents (SD) | | |
|---|---|---|---|---|---|---|---|
| | | Rouge-1 | Rouge-2 | Rouge-L | Rouge-1 | Rouge-2 | Rouge-L |
| BOW | MMR | 0.362 | 0.238 | 0.312 | 0.369 | 0.218 | 0.317 |
| | xDTD | 0.376 | 0.249 | 0.317 | 0.344 | 0.197 | 0.293 |
| | J-xDTD | 0.387 | 0.264 | 0.327 | 0.349 | 0.203 | 0.298 |
| BOW+DM | MMR | 0.406 | 0.290 | 0.357 | 0.388 | 0.241 | 0.339 |
| | xDTD | 0.443 | 0.331 | 0.392 | 0.385 | 0.248 | 0.339 |
| | J-xDTD | 0.445 | 0.328 | 0.395 | 0.385 | 0.248 | 0.339 |
| BOW+DBOW | MMR | 0.418 | 0.293 | 0.364 | 0.395 | 0.246 | 0.347 |
| | xDTD | 0.415 | 0.304 | 0.369 | 0.371 | 0.236 | 0.329 |
| | J-xDTD | 0.410 | 0.300 | 0.363 | 0.371 | 0.236 | 0.329 |

**Table 3.** *Summarization results achieved by several well-studied or/and state-of-the-art unsupervised methods.*

| Method | Text Documents (TD) | | | Spoken Documents (SD) | | |
|---|---|---|---|---|---|---|
| | Rouge-1 | Rouge-2 | Rouge-L | Rouge-1 | Rouge-2 | Rouge-L |
| LSA | 0.362 | 0.233 | 0.316 | 0.345 | 0.201 | 0.301 |
| CBOW | 0.369 | 0.224 | 0.308 | 0.365 | 0.206 | 0.313 |
| SG | 0.367 | 0.230 | 0.306 | 0.358 | 0.205 | 0.303 |
| GloVe | 0.367 | 0.231 | 0.308 | 0.364 | 0.214 | 0.312 |
| ULM | 0.411 | 0.298 | 0.361 | 0.364 | 0.210 | 0.307 |
| MRW | 0.412 | 0.282 | 0.358 | 0.332 | 0.191 | 0.291 |
| LexRank | 0.413 | 0.309 | 0.363 | 0.305 | 0.146 | 0.254 |
| SM | 0.414 | 0.286 | 0.363 | 0.332 | 0.204 | 0.303 |
| ILP | 0.442 | 0.337 | 0.401 | 0.348 | 0.209 | 0.306 |

## 7. CONCLUSIONS & FUTURE WORK

In this paper, two novel coverage-based methods have been proposed and extensively evaluated for extractive SDS. In addition, several document and sentence representation methods have also been compared in this study. Finally, these methods have been further integrated into a formal summarization framework. Experimental results demonstrate the effectiveness of the proposed coverage-based methods in relation to several state-of-the-art baselines compared in the paper, thereby indicating the potential of such a new summarization framework. For future work, we will explore other feasible ways to enrich the representations of documents/sentences and integrate extra cues, such as speaker identities or prosodic (emotional) information, into the proposed framework. We also plan to investigate more elegant and robust techniques to estimate sub-themes of a given document. Furthermore, how to accurately estimate the component models involved in the proposed methods will be one of the interesting research directions.